\documentclass[10pt,twocolumn,letterpaper]{article}

\usepackage{iccv}
\usepackage{times}
\usepackage{epsfig}
\usepackage{graphicx}
\usepackage{amsmath}
\usepackage{amssymb}
\usepackage{enumitem}
\usepackage{booktabs}
\usepackage{multirow}
\usepackage[normalem]{ulem}
\usepackage[table,xcdraw]{xcolor}

\usepackage[pagebackref=true,breaklinks=true,letterpaper=true,colorlinks,bookmarks=false]{hyperref}

\iccvfinalcopy 

\def\iccvPaperID{8774} 

\ificcvfinal\fi

\newcolumntype{x}[1]{>{\centering\arraybackslash}p{#1pt}}
\newcolumntype{y}[1]{>{\raggedright\arraybackslash}p{#1pt}}
\newcolumntype{z}[1]{>{\raggedleft\arraybackslash}p{#1pt}}

\definecolor{baselinecolor}{gray}{.9}
\newcommand{\baseline}[1]{\cellcolor{baselinecolor}{#1}}
\useunder{\uline}{\ul}{}
\newcommand{\tablestyle}[2]{\setlength{\tabcolsep}{#1}\renewcommand{\arraystretch}{#2}\centering\footnotesize}
\newcommand{\firstparagraph}[1]{\noindent\textbf{#1}}
\renewcommand{\paragraph}[1]{\vspace{1.25mm}\noindent\textbf{#1}}

\begin{document}

\title{Forecast-MAE: Self-supervised Pre-training for Motion Forecasting with Masked Autoencoders}

\author{
	Jie Cheng$^{1}$ \quad Xiaodong Mei$^{1}$ \quad Ming Liu$^{1,2}$\thanks{Corresponding author: \textit{Ming Liu}} \\
	HKUST$^{1}$ \quad HKUST(GZ)$^{2}$\\
	{\tt\small \{jchengai, xmeiab\}@connect.ust.hk,\quad eelium@ust.hk}
}

\maketitle
\ificcvfinal\thispagestyle{empty}\fi

\begin{abstract}

    This study explores the application of self-supervised learning (SSL) to the task of motion forecasting, an area that has not yet been extensively investigated despite the widespread success of SSL in computer vision and natural language processing. To address this gap, we introduce Forecast-MAE, an extension of the mask autoencoders framework that is specifically designed for self-supervised learning of the motion forecasting task. Our approach includes a novel masking strategy that leverages the strong interconnections between agents’ trajectories and road networks, involving complementary masking of agents’ future or history trajectories and random masking of lane segments. Our experiments on the challenging Argoverse 2 motion forecasting benchmark show that Forecast-MAE, which utilizes standard Transformer blocks with minimal inductive bias, achieves competitive performance compared to state-of-the-art methods that rely on supervised learning and sophisticated designs. Moreover, it outperforms the previous self-supervised learning method by a significant margin. 
    Code is available at \url{https://github.com/jchengai/forecast-mae}. 
\end{abstract}

\begin{figure}[t]
\begin{center}
\includegraphics[width=1.0\linewidth, trim=0 0 10 0,clip]{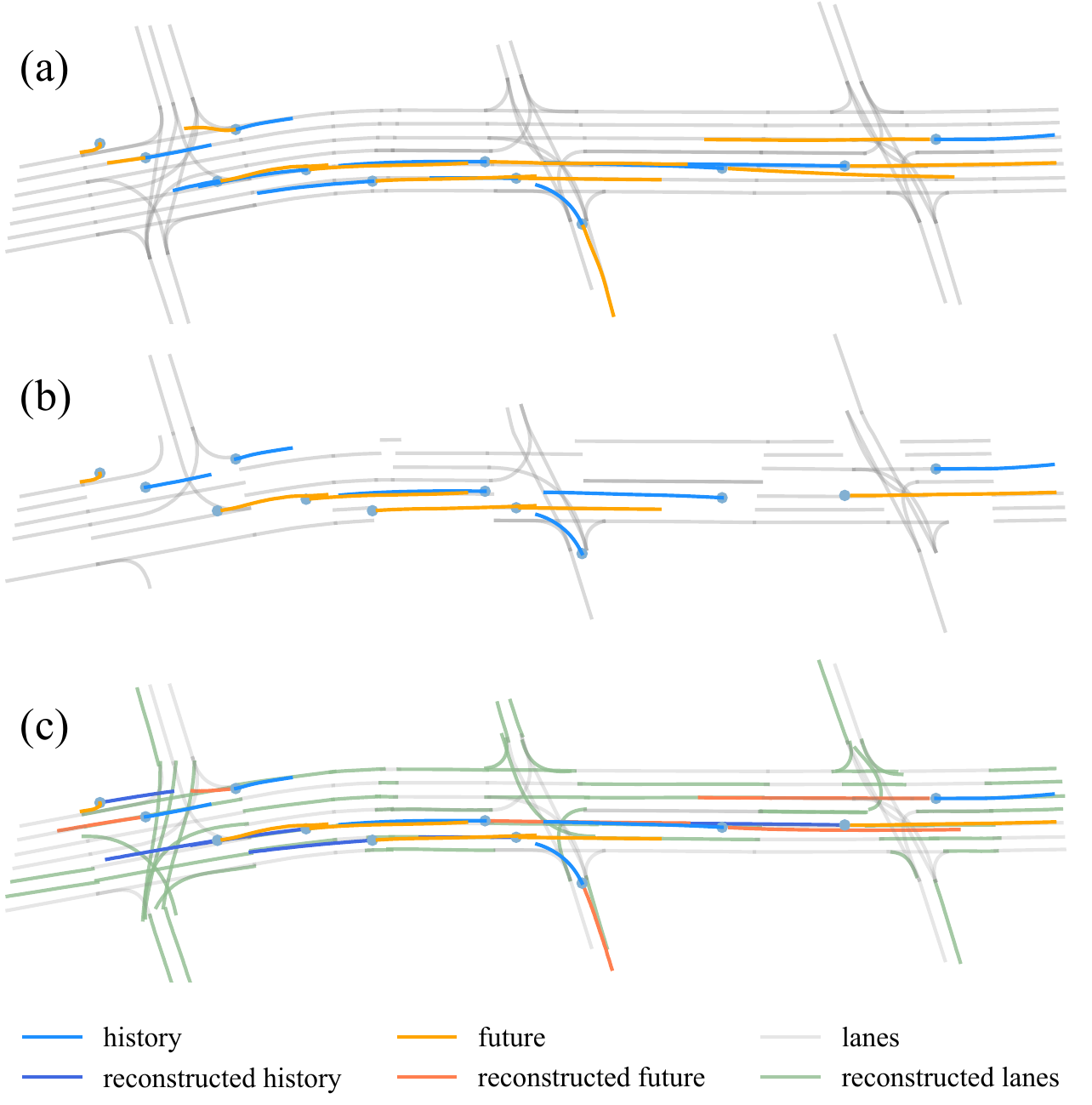}
\end{center}
\vspace{-1em}
   \caption{
   Reconstruction result on Argoverse 2 \textit{validation} scenario. 
   (a) The origin scenario. 
   (b) 50\% of agents' trajectory is masked using a complementary masking strategy (either history or future is masked). 50\% of the lane segments are masked randomly. 
   (c) Scenario reconstructed by the proposed Forecast-MAE.
}
\label{fig:reconsturction}
\vspace{-1.2em}
\end{figure}

\section{Introduction}
Motion forecasting is a rapidly developing research field that plays a critical role in advanced autonomous driving systems~\cite{huang2022survey}. This task involves predicting the future trajectories of other vehicles and pedestrians, while taking into account the intricate interactions and road layouts. The inherent multi-modal driving behaviors of agents, combined with diverse road networks, make motion forecasting an especially challenging undertaking.

Self-supervised learning (SSL) is an innovative approach that enables the acquisition of valuable latent features from unlabelled data. By pre-training the model on pretext tasks and pseudo-labels derived from the data, and subsequently fine-tuning on downstream tasks, SSL has demonstrated an ability to learn more extensive and adaptable latent features, leading to remarkable advancements in computer vision~\cite{Caron_2021_ICCV} and natural language processing (NLP)~\cite{devlin2018bert}. Nevertheless, despite its widespread popularity and success, there remains a notable lack of exploration of SSL in the motion forecasting domain. We have identified two principal challenges associated with integrating SSL into motion forecasting:

(i) 
 Motion forecasting pre-training requires annotated data, which sets it apart from fields such as computer vision and NLP where unlabeled raw inputs are easily accessible. In motion forecasting, we rely on annotated tracking sequences and hand-crafted high-definition maps that are typically collected by expensive onboard sensors and require human annotation labor~\cite{chang2019argoverse, wilson2argoverse, ettinger2021large}. This poses a challenge to scaling up self-supervised pre-training, a key aspect of SSL's success. To address this challenge, very recent work PreTraM~\cite{xu2022pretram} proposed generating additional rasterized map patches (28.8M) cropped from local regions of the entire HD map to train a robust map encoder with contrastive learning. Although this approach yielded notable performance improvements compared to the baseline, it is limited to models based on rasterized map representations, which have a significant performance gap compared to more recent vector-based or graph-based models. However, another pioneering work, SSL-Lanes~\cite{bhattacharyyassl}, demonstrated that carefully designed pretext tasks can significantly enhance performance without using extra data by learning richer features. In this paper, we follow this approach to learn better and more generalized features using the existing dataset.

(ii)
The task of motion forecasting involves incorporating multiple modal inputs, such as static map features, spatiotemporal agent motion features, and semantic scene contexts~\cite{tang2019multiple,liang2020learning, gao2020vectornet, chai2020multipath, shimotion, zhao2021tnt, liu2021multimodal, deo2022multimodal, zhou2022hivt}. While various self-supervised learning methods have proven successful in dealing with single-modal inputs such as the image~\cite{Caron_2021_ICCV}, text~\cite{devlin2018bert}, or point cloud~\cite{yu2022point, pang2022masked}, developing pretexts that establish cross-modal interconnections is not an easy task.
SSL-Lanes concentrated on designing pretext tasks for each specific input modality, such as lane node masking or agent maneuver classification. Nevertheless, they did not explore the combination of these different tasks or develop pretext tasks that explicitly involve multiple modal inputs.
Authors of PreTraM drew inspiration from CLIP's~\cite{radford2021learning} cross-modal contrastive learning framework involving text and images. They devised a technique for pre-training map and trajectory encoders by pairing batches of (map, trajectory) training instances. Nevertheless, their approach merely encompasses the history trajectory-map connection, thereby restricting the scope of modality interconnections to a particular type.
This study confronts this challenge by utilizing a masked autoencoder framework that can assimilate all cross-modal interdependencies within a unified scene reconstruction task.

The masked autoencoder (MAE)~\cite{he2022masked} has garnered significant attention due to its recent achievements in image-based self-supervised learning. This approach involves masking a portion of the input data and reconstructing the missing part using an autoencoder structure. The effectiveness of MAE has also been demonstrated in other domains, such as audio~\cite{huangmasked} and point cloud~\cite{pang2022masked}. An intriguing question arises: \textit{can we extend MAE to motion forecasting?} Indeed, motion forecasting itself can be viewed as a masking and reconstructing task, wherein the future trajectory of agents is masked and predicted. Based on the strong correlation between agents' historical and future trajectories and road networks, we further extend this concept to the entire scene reconstruction. 
Specifically, we mask agents' history trajectory or future trajectory in a complementary manner (\ie either history or future is masked), and randomly mask non-overlapping lane segments, shown in Figure \ref{fig:reconsturction}. 
This masking scheme offers several advantages. Firstly, the model must learn how to reconstruct the future from past motion and, in turn, infer history from the future, with limited access to lane structures. This pretext task allows the model to establish a robust bidirectional relationship between past and future motion. Secondly, the model learns to reconstruct lane segments by jointly utilizing neighboring visible lanes, agents' history and future trajectories, thereby establishing a more profound cross-modal understanding.

To this end, we introduce Forecast-MAE, an extension of the masked autoencoder framework specifically designed for self-supervised learning of the motion forecasting task. 
Our methodology comprises a novel masking design that exploits the strong interdependencies among all agents' trajectories and road networks. Despite being simple and incorporating minimal inductive bias, our proposed Forecast-MAE performs strongly on the challenging Argoverse 2 (AV2) motion forecasting benchmark~\cite{wilson2argoverse} and significantly outperforms the previous self-supervised learning method.

Our contribution can  be summarized as follows:
\begin{itemize}[noitemsep,nolistsep]
    \item To our best knowledge, we propose the first masked autoencoding framework for self-supervised learning on the motion forecasting task. 
    Without extra data or pseudo-labels, our method greatly improves the performance of motion forecasting through pre-training compared to training from scratch.
    \item We introduce a straightforward yet highly effective masking scheme that facilitates the learning of bi-directional motion connections and cross-modal relationships within a single reconstruction pretext task.   
    \item We show that our approach, based entirely on standard Transformers with minimal inductive bias, achieves competitive performance compared to the state-of-the-art with supervised learning on the challenging Argoverse 2 benchmark, and significantly outperforms the previous self-supervised learning method. 
    \item Our findings suggest that SSL can be a promising approach for motion forecasting, and we anticipate that this may spark greater interest in the field. 
\end{itemize}

\section{Related Work}

\firstparagraph{Motion Forecasting}.
The performance of motion forecasting models has significantly advanced in recent years, primarily attributable to the amplified interest in self-driving vehicles and the widespread availability of standard benchmarks. 
Herein, we concisely outline three key aspects contributing to its improvements.

(i) \textit{Improvement on scene representation}. 
In the early stages, rasterized top-down semantic images are commonly utilized for scene representation, and off-the-shelf image encoders are used for learning~\cite{tang2019multiple, phan2020covernet, chai2020multipath, gilles2021home}. Although this image-based representation is simple and unified, it inevitably results in the loss of detailed structural information during rasterization. The popularity of vectorized representations has increased significantly with the introduction of VectorNet~\cite{gao2020vectornet}, owing to their higher representation capacity and significantly stronger performance. Furthermore, graphs~\cite{liang2020learning, zeng2021lanercnn, deo2022multimodal, gilles2022gohome, khandelwal2020if} are widely used as another promising scene representation. TPCN~\cite{ye2021tpcn}, as a standalone approach, achieves impressive results by treating the agents' trajectories and lanes as the point cloud.

(ii) \textit{Improvement on model architectures}. 
Early rasterized methods naturally relied on well-established convolutional networks. Later, inspired by the impressive performance of Transformer~\cite{vaswani2017attention}, attention mechanisms have been extensively used for interaction modeling and information aggregation, given their superior flexibility and efficacy. Some works~\cite{ngiam2022scene,liu2021multimodal, zhou2022hivt, girgislatent} have directly incorporated transformers for forecasting and achieved satisfactory outcomes. A more recent work, MTR~\cite{shimotion}, builds on cutting-edge vision object detection architecture DETR~\cite{carion2020end}, resulting in state-of-the-art performance.
Advances in the graph neural network (GNN) domain are also widely explored~\cite{zeng2021lanercnn, gilles2022gohome, khandelwal2020if, deo2022multimodal,9981037, 9560908, chen2020comogcn}. 
LaneGCN~\cite{liang2020learning} modified graph convolutional operation tailed for lane graph encoding. 
HDGT~\cite{jia2022hdgt} uses the heterogeneous graph to encode different types of agents and map elements. 
HiVT~\cite{zhou2022hivt}, QCNet~\cite{zhou2023query} and~\cite{jia2022multi} explores different corrdiantes systems.

(iii) \textit{Introducing of prior knowledge}. 
Incorporating prior knowledge to tackle the complex problem of multi-modal future prediction has become increasingly prevalent in recent literature. Several works utilize predefined candidate trajectories~\cite{phan2020covernet,song2022learning} or anchor points~\cite{chai2020multipath, varadarajan2022multipath++} by clustering the ground truth or generating with planners. Another line of research involves sampling goals within the drivable areas and utilizing a two-stage prediction pipeline~\cite{ zeng2021lanercnn, deo2022multimodal, zhao2021tnt, gu2021densetnt, gilles2021home, gilles2022gohome}. DCMS~\cite{ye2022dcms} introduces temporal consistent constraints based on the assumption that predictions should not change abruptly. However, these methods typically require additional computation or have a higher model complexity.

Despite significant advancements in motion forecasting, there is a recent trend towards greater architectural complexity and utilization of prior knowledge. In this study, we explore a different direction for enhancing performance, namely self-supervised learning. By leveraging the simplicity of the MAE framework, we demonstrate that our proposed Forecast-MAE, employing a standard transformer architecture with minimal prior knowledge, can achieve competitive performance compared to state-of-the-art supervised learning-based methods with sophisticated designs.

\paragraph{Self-supervised Learning in Motion Forecasting}. 
There are only a few studies that explore SSL in motion forecasting. To the best of our knowledge, VectorNet is the earliest work that incorporates a BERT-like~\cite{devlin2018bert} graph completion task to better capture interactions between agents and maps. However, it is a very preliminary attempt, and the graph completion is treated as an auxiliary training objective that is jointly optimized with the motion forecasting task. PreTraM and SSL-Lanes are two recent works that systematically study SSL.
The authors of PreTraM believe that the scarcity of trajectory data restricts the application of SSL in motion forecasting. They generate additional local map patches from the entire maps and leverage single-modal and cross-modal contrastive learning to pre-train the map and trajectory encoders separately.
In contrast, our method adopts a completely different MAE-based framework, where representations of different modalities are learned jointly. 
SSL-Lanes demonstrated that SSL could learn better latent features without using extra data. 
It studied four pretext tasks, each focusing on one specific input modality, such as lane masking or agents' maneuver classification. However, they do not explore combining these different tasks or designing pretext tasks involving multi-modal inputs.
On the contrary, the proposed Forecast-MAE learns cross-modal interconnections by design and outperforms SSL-Lanes by a large margin.

\begin{figure*}[hbt!]
\begin{center}
\includegraphics[width=0.95\linewidth]{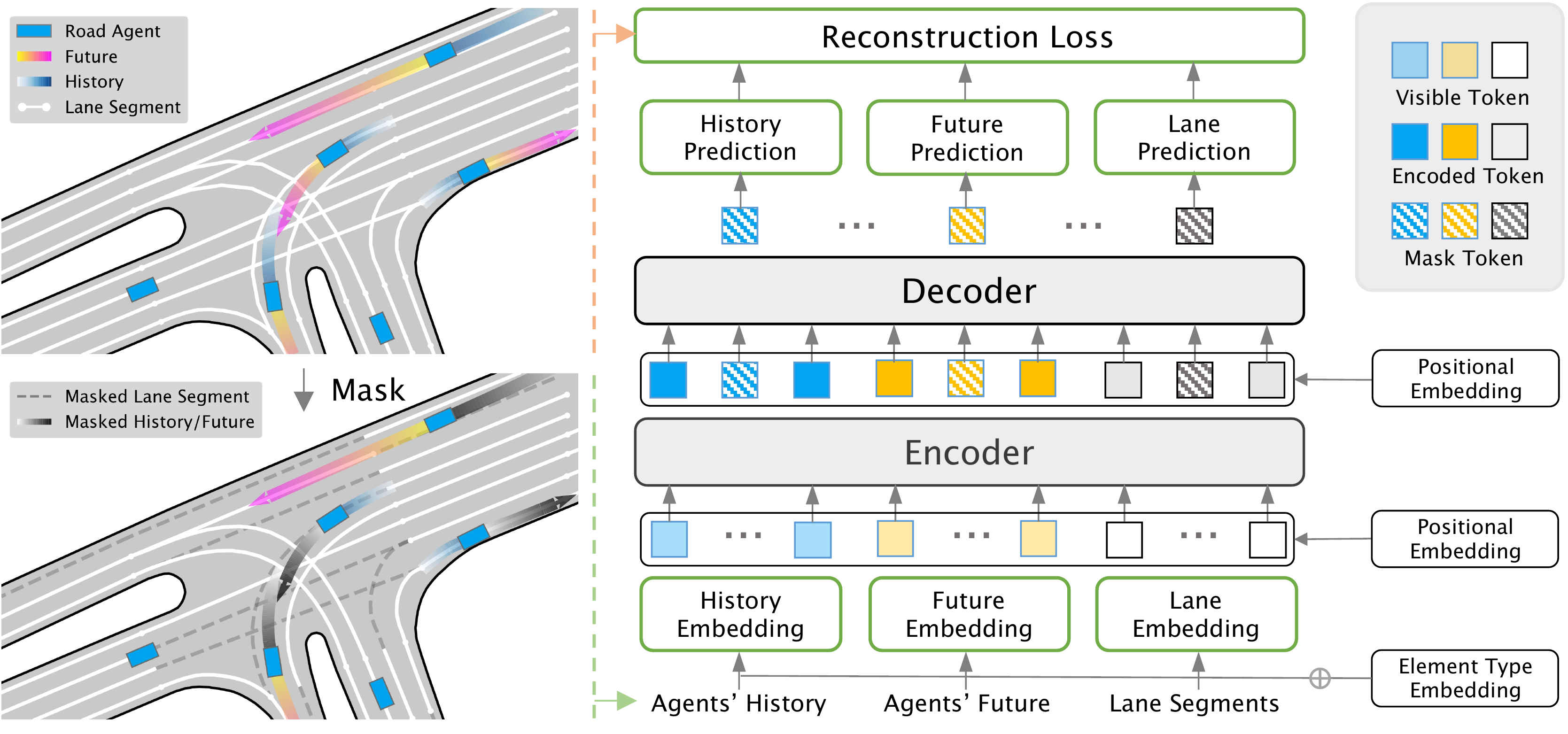}
\vspace{-0.7em}
\end{center}
   \caption{
   \textbf{Overall pre-training scheme of our Forecast-MAE}. 
   The left part shows the masking process of an example scenario (two agents are static within the observation horizon). We randomly mask out the entire agents' history or future trajectory, as well as lane segments. 
   The pre-training scheme is shown on the right. 
   Only the visible history, future trajectory, and lane segments will be embedded into tokens and processed by the encoder. 
   Three different types of mask tokens are added to the input sequence of the decoder to reconstruct history, future trajectory and lane segment, respectively. 
   }
\label{fig:Forecast-MAE}
\vspace{-1em}
\end{figure*}

\section{Methodology}
We propose Forecast-MAE, a simple and neat MAE-based framework for self-supervised pre-training of the motion forecasting task. 
The pre-training process is illustrated in Figure \ref{fig:Forecast-MAE}.
Visible agents' history/future trajectories and lane segments are embedded as tokens and then processed with a standard transformer encoder. 
Following the asymmetric design of the vision MAE~\cite{he2022masked}, different mask tokens are added to the decoder's input sequence and later used to reconstruct the masked trajectories and lane segments with simple prediction heads.

\subsection{Masking}
In contrast to all current self-supervised learning frameworks for motion forecasting, we utilize the future trajectories of agents as an additional input for pre-training. Our experiments reveal that masking future trajectories is a crucial aspect for Forecast-MAE to be effective. 
To begin, the road maps are initially segmented into non-overlapping lane sections. We then randomly mask a subset of lane segments according to a uniform distribution. 
The masking technique for agents differs slightly. Although random masking is still employed for agent trajectories, we only mask the history or future trajectory of each agent (\eg, 40\% of agents retain their history, while the remaining 60\% retain their future). We refer to this process as \textit{complementary random masking}. 
This constraint is sensible, as reconstructing trajectories from a single pose is not a meaningful pretext task when both history and future trajectories are masked.

\subsection{Input Representation and Embedding}
Following the popular vectorized representation, we treat all agents' trajectories and lane segments as polylines. 
Specifically, we denote the history trajectories of $N$ agents as $A_{H} \in \mathbb{R}^{N\times T_H \times C_H}$, where $T_H$ is the number of history frames and $C_H$ is the history feature channels including step-wise displacement/velocity difference, and a padding flag indicating the observation status of this frame. 
Similarly, the future trajectories are denoted as $A_F \in \mathbb{R}^{N \times T_F \times C_F}$, where $T_F$ is the number of future frames, $C_F$ is the future feature channels, including future coordinates normalized to the current position of agents and a padding flag indicating availability. 
The non-overlapping lane segments are denoted as $L \in \mathbb{R}^{M \times P \times C_L}$, where $M$ is the number of lane segments within a certain radius of the target agent, $P$ is the number of points of each polyline and $C_L$ is the lane feature channels (e.g., coordinates, availability). 
Note that we normalize all the coordinates of each lane polyline to its geometric center. 

The primary goal of the embedding layer is to encode sequential features into one-dimensional vectors or tokens that can be directly processed by the standard Transformer.
We use a Feature Pyramid Network (FPN)~\cite{lin2017feature} similar to LaneGCN to fuse multi-scale agent motion features. 
1D neighborhood attention~\cite{hassani2022neighborhood} is employed at each scale to extract local motion features. 

Agents' historical and future features are embedded separately. To capture a broader range of the road map, we employ a lightweight mini-PointNet~\cite{qi2017pointnet}, mainly comprising MLPs and max pooling layers, to embed lane polylines.
The embedding process can be formulated as
\begin{align}
T_H = \text{FPN}(A_H),\;T_F = \text{FPN}(A_F)&,\; T_{H,F} \in \mathbb{R}^{N \times C} \\
T_L = \text{MiniPointNet}(L)&,\; T_L \in \mathbb{R}^{M \times C},
\end{align}
where $T_H, T_F, T_L$ are history, future, lane tokens respectively, $C$ is the embedding dimension. 

Semantic attributions such as agent category (e.g., vehicle, pedestrian, cyclist) or lane types are initialized as learnable embeddings and added to the embedded tokens. 
Given that the coordinates of agents and lane features are normalized, it is crucial to include global position information in the tokens. 
The position embedding (PE) is implemented with a simple two-layer MLP following~\cite{yu2022point}, formulated as 
\begin{align}
    \text{PE} = \text{MLP}\big(\left[x, y, \cos(\theta), \sin(\theta \right]\big), \text{PE} \in \mathbb{R}^C 
\end{align}
where $(x,y,\theta)$ is the latest observed pose of agents or the geometric center pose for lane polylines. 
The PE is added to the tokens before being processed by the autoencoder.

\subsection{AutoEncoder}
The autoencoder is entirely based on standard Transformers. 
The encoder consists of several Transformer blocks and only encodes concatenated visible agents and lane tokens, resulting in encoded latent tokens $T_E \in \mathbb{R}^{(N + M)\times C}$. 

Following the asymmetric autoencoder design of MAE, history, future, and lane mask tokens $M=(M_H,M_F,M_L)$ are added together with the encoded latent tokens as the input sequence of the decoder and then output the decoded mask tokens $M'=(M_H^{'}, M_F^{'},M_L^{'})$ after the decoding. 
Positional embeddings are added to the full input sequence, including the mask tokens.
Each type of mask token is a learned vector shared by the corresponding type of masked element. 
The autoencoding process is formulated as
\begin{align}
    T_E &= \text{Encoder}\big(\text{concat}\left(T_H, T_F, T_L\right) + \text{PE}\big), \\ 
    M' &= \text{Decoder}\big(\text{concat}\left(T_E,M\right)+\text{PE}\big).
\end{align}
The decoded mask tokens are subsequently used for reconstructing the masked element through a simple prediction head, which is implemented as a linear projection layer in practice.

\subsection{Reconstruction Target}
The prediction heads predict the normalized 2-dimensional coordinates of history/future trajectories $P_{H/F}$ and lane polylines $P_L$,
\begin{align}
    P_H &= \text{PredictionHead}(M_H'), \;P_H \in \mathbb{R}^{\alpha N \times T_H \times 2}, \\
    P_F &= \text{PredictionHead}(M_F'), \;P_F \in \mathbb{R}^{(1-\alpha) N \times T_F \times 2},\\
    P_L &= \text{PredictionHead}(M_L'), \;P_L \in \mathbb{R}^{\beta M \times P \times 2},
\end{align}
where $\alpha$ is the agents' history mask ratio, $\beta$ is the lane segments mask ratio. 
We use L1 loss $\mathcal{L}_H, \mathcal{L}_F$ for trajectory reconstruction and mean squared error (MSE) loss $\mathcal{L}_L$ for lane polyline reconstruction, and $w_H, w_F, w_L$ correspond to the loss weight respectively. The final loss is 
\begin{align}
\begin{split}
    \mathcal{L}_{MAE} = w_H \mathcal{L}_H + w_F \mathcal{L}_F + w_L \mathcal{L}_L.
\end{split}
\end{align}

\subsection{Motion Forecasting}
For the target motion forecasting task, we adopt an end-to-end fine-tuning approach.
During fine-tuning, the following modifications are made to the pre-training model: (1) we discard the MAE decoder and mask tokens; (2) agents' future features are eliminated from the input, and masking is not employed; (3) the pretext prediction heads are substituted with a multi-modal future decoder.

\paragraph{Multi-modal decoder.} 
Given the multi-modal nature of agents' behavior, motion forecasting entails producing multiple potential future predictions, distinct from the masked future reconstruction pretext task.
To maintain a neat framework with minimal inductive bias, we implement the multi-modal decoder using a simple three-layer MLP. A separate three-layer MLP is utilized to generate the confidence score for each prediction.
The decoding process can be formulated as 
\begin{align}
    P_{Traj} = \text{MLP}(T_H'),\; P_{Traj} \in \mathbb{R}^{N\times K \times T_F \times 2},
\end{align}
where $T_H'$ is the encoded history tokens and $K$ is the number of output modes. 
The predicted future trajectories are normalized to each agent's latest observed position. 

\paragraph{Training loss.} 
We adopt the widely used Huber loss for trajectory regression and cross-entropy loss for confidence classification with equal weights. 
The winner-take-all strategy is employed, which only optimizes the best prediction with minimal average prediction error to the ground truth. 
We compute the loss using all agents present in the scene. 

\section{Experiments}

\subsection{Experimental Setup}
\paragraph{Dataset.} 
We evaluate the proposed framework on the recently released large-scale Argoverse 2 (AV2) dataset. This dataset includes 250K non-overlapping scenarios, divided into 199,908, 24,988, and 24,984 samples for training, validation, and testing, respectively. Each sample contains 5 seconds of history and requires a prediction of 6 seconds in the future, with a sampling rate of 10 Hz.
Every scenario includes a focal track agent that needs to be predicted, with detailed high-definition map patches provided for each sample.
We choose to evaluate on the Argoverse 2 dataset as it offers the best balance between diversity and dataset size. The popular Argoverse 1~\cite{chang2019argoverse} dataset has a similar size but lacks scenario diversity (e.g., the majority of vehicles driving straight-forward). Conversely, Argoverse 2 is intended to be more varied and complicated. 
Another widely-used dataset, the Waymo Open Motion Dataset (WOMD)~\cite{ettinger2021large}, has similar scenario complexity but only contains less than half the number of scenarios (104K). We believe that a larger and more complex dataset is more appropriate for evaluating SSL frameworks. 

\paragraph{Metrics.}
We use the official benchmark metrics, including minADE, minFDE, MR, and brier-minFDE, which refer to six prediction modes, if not specified. 

\paragraph{Implementation Details.}
Detailed model architecture and training settings are provided in the supplementary.

\begin{figure}[t]\centering
\begin{center}
\includegraphics[width=1.0\linewidth,trim=0 0 5 0, clip]{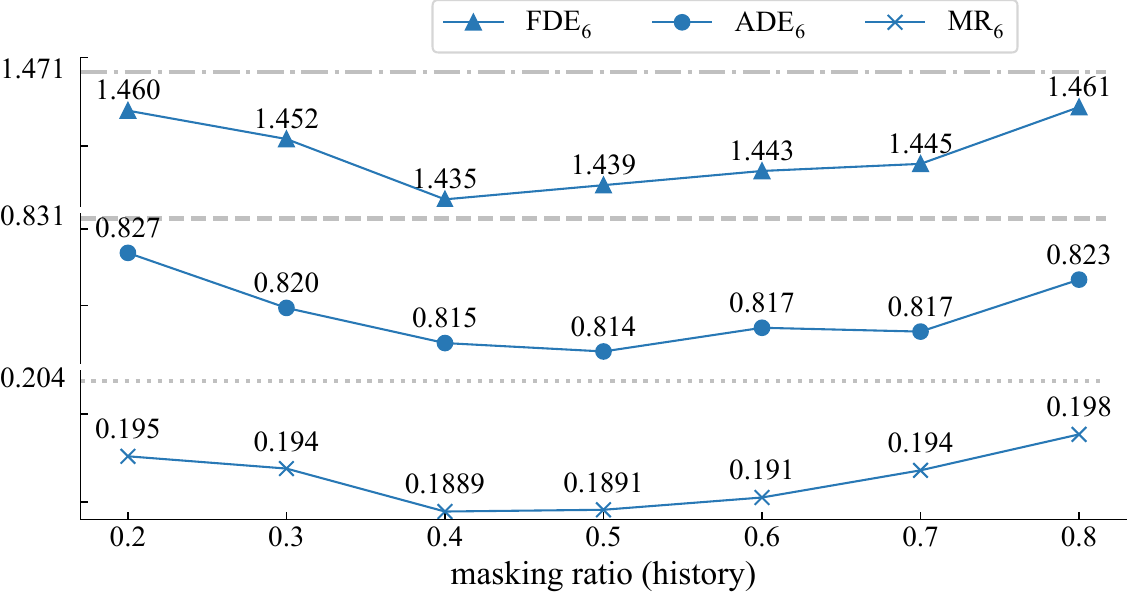} \\
 \hspace{0.1em}
\includegraphics[width=1.0\linewidth,trim=0 0 5 0, clip]{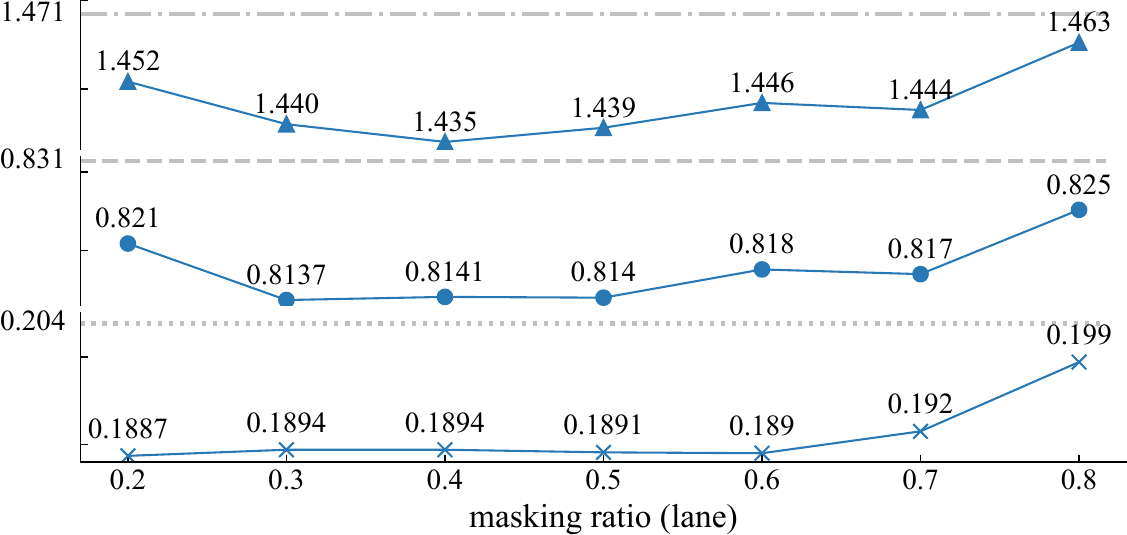}
\end{center}
\vspace{-0.5em}
   \caption{\textbf{The impact of the history and lane masking ratio}. 
   The marked blue lines show the fine-tuning results on different metrics (lower is better). 
   The dashed lines in grey correspond to the results of the baseline model (trained from scratch for 30 epochs).
   }
\label{fig:ab-mask_ratio}
\vspace{-.8em}
\end{figure}

\subsection{Ablation Study}
We conduct ablation studies on the Argoverse 2 validation set. 
By default, the pre-training epoch is set to 40, the fine-tuning epoch to 30, the history and lane mask ratio to 0.5, and the encoder and decoder depth to 4. 
The pre-training is only conducted on the training set. 

\paragraph{Masking ratio.} 
Figure \ref{fig:ab-mask_ratio} depicts the impact of varying masking ratios. Employing a well-balanced masking ratio, ranging from 40\% to 50\%, between an agent's history and future leads to the most favorable outcomes, in agreement with common sense. We posit that a balanced masking ratio for agent trajectory helps prevent the learning of biased features by the model and enhances its comprehension of the bidirectional relationship between historical and future motion. This is further demonstrated by the performance of extreme history masking ratios (20\% and 80\%), which significantly underperform.

Forecast-MAE is relatively insensitive to lane masking ratio, as a wide range of ratios (30\% to 60\%) perform well. 
Nonetheless, when the lane masking ratio exceeds 70\%, performance suffers notably. 
The possible reason is with a masking ratio of more than 70\%, most of the road structure information loses, which tremendously increases the difficulty of scene reconstruction and geometry feature extraction from the map. 
Conversely, when the lane masking ratio is below 20\%, both ADE and FDE experience a significant increase. We deduce that the masked lanes can be easily extrapolated by nearby visible lanes when only a small subset of lanes are masked. 

\begin{table}[]
\tablestyle{8pt}{1.1}
\begin{tabular}{x{12}x{12}x{12}x{12}x{20}x{20}x{20}}
\toprule
\begin{tabular}[c]{@{}c@{}}Hist.\\ mask\end{tabular} & \begin{tabular}[c]{@{}c@{}}Lane\\ mask\end{tabular} & \begin{tabular}[c]{@{}c@{}}Fut.\\ input\end{tabular} & \multicolumn{1}{c}{\begin{tabular}[c]{@{}c@{}}Fut.\\ mask\end{tabular}} & minADE & minFDE & MR \\ \midrule
 & \checkmark &  &  & 0.865 & 1.51 & 0.212 \\
\multicolumn{1}{l}{} & \checkmark & \checkmark &  & 0.828 & 1.47 & 0.203 \\
\checkmark & \multicolumn{1}{l}{} & \checkmark &  & 0.864 & 1.53 & 0.216 \\
\checkmark & \checkmark & \checkmark &  & 0.866 & 1.52 & 0.214 \\
\multicolumn{1}{l}{} & \checkmark & \checkmark & \multicolumn{1}{c}{\checkmark} & 0.820 & 1.45 & 0.198 \\
\baseline{\checkmark} & \baseline{\checkmark} & \baseline{\checkmark} & \multicolumn{1}{c}{\baseline{\checkmark}} & \baseline{\textbf{0.814}} & \baseline{\textbf{1.44}} & \baseline{\textbf{0.189}} \\ \midrule
\multicolumn{4}{c}{training from scratch} & 0.8314 & 1.471 & 0.2038 \\ \bottomrule
\end{tabular}
\vspace{.8em}
\caption{
\textbf{Results of different input and masking strategies.} 
History and lane are used as input for experiments.  
When history and future trajectories are simultaneously masked, it is masked in a complementary manner. 
For other situations, we use random masking by default. 
}
\label{tab:masking_strategy}
\vspace{-0.5em}
\end{table}

\begin{table}[t]
\tablestyle{8pt}{1.1}
\begin{tabular}{x{45}x{40}x{40}x{40}}
\toprule
encoder depth & minADE & minFDE & MR \\ \midrule
2 & 0.854 & 1.55 & 0.221 \\
3 & 0.823 & 1.46 & 0.198 \\
\baseline{4} & \baseline{\textbf{0.814}} & \baseline{1.44} & \baseline{0.189} \\
5 & 0.815 & \textbf{1.43} & \textbf{0.188} \\ \bottomrule
\end{tabular}
\vspace{.8em}
\caption{\textbf{Results of different encoder depth.} A encoder depth of 4 offers the best performance-efficiency trade-off.}
\label{tab:encoder_depth}
\vspace{-1em}
\end{table}

\begin{table*}[t]
\centering
\def\arraystretch{1.1} 
\resizebox{0.95\textwidth}{!}{
\begin{tabular}{y{60}y{100}x{40}x{40}x{40}x{40}x{40}x{40}x{40}}
\toprule
\multicolumn{1}{l}{} & Method & minADE$_1$ & minFDE$_1$ & MR$_1$ & minADE$_6$ & minFDE$_6$ & MR$_6$ & b-FDE$_6$ \\ \midrule
 & THOMAS~\cite{gillesthomas} & 1.96 & 4.71 & 0.64 & 0.88 & 1.51 & 0.20 & 2.16 \\
 & GoReLa~\cite{cui2022gorela} & 1.82 & 4.62 & 0.61 & 0.76 & 1.48 & 0.22 & 2.01 \\
 & GANet~\cite{wang2022ganet} & 1.78 & 4.48 & {\ul 0.60} & 0.73 & {\ul 1.35} & \textbf{0.17} & 1.97 \\
 & QML w/ ensemble~\cite{su2022qml}  & 1.84 & 4.98 & \textbf{0.59} & \textbf{0.69} & 1.39 & 0.19 & 1.95 \\
\multirow{-4}{*}{\begin{tabular}[c]{@{}c@{}}Supervised \\ Learning\end{tabular}} & BANet w/ ensemble~\cite{zhangbanet} & 1.79 & 4.61 & {\ul 0.60} & {\ul 0.71} & 1.36 & 0.19 & {\ul 1.92} \\ \midrule
 & Lane Masking & 2.167 & 5.675 & 0.671 & 0.835 & 1.698 & 0.248 & 2.379 \\
 & Dist. to Inter. & 2.176 & 5.71 & 0.667 & 0.839 & 1.710 & 0.248 & 2.391 \\
\multirow{-3}{*}{SSL-Lanes~\cite{bhattacharyyassl}} & S/F Classification & 2.218 & 5.905 & 0.687 & 0.828 & 1.671 & 0.249 & 2.352 \\ \cmidrule(lr){2-2}
 & Scratch & 1.845 & 4.602 & 0.623 & 0.727 & 1.427 & 0.187 & 2.062 \\
 & Fine-tune w/o ensemble & {\ul 1.741} & {\ul 4.355} & { 0.607} & {\ul 0.709} & 1.392 & {\ul 0.172} & 2.029 \\
\multirow{-3}{*}{Forecast-MAE} & \cellcolor[HTML]{EFEFEF}Fine-tune w/ ensemble & \cellcolor[HTML]{EFEFEF}\textbf{1.658} & \cellcolor[HTML]{EFEFEF}\textbf{4.145} & \cellcolor[HTML]{EFEFEF}\textbf{0.592} & \cellcolor[HTML]{EFEFEF}\textbf{0.690} & \cellcolor[HTML]{EFEFEF}\textbf{1.338} & \cellcolor[HTML]{EFEFEF}0.173 & \cellcolor[HTML]{EFEFEF}\textbf{1.911} \\ \bottomrule\bottomrule
\vspace{0.1em}
 & Lane Masking & 2.014 & 5.194 & 0.649 & 0.850 & 1.520 & 0.220 & 2.197 \\
 & Dist. to Inter. & 2.006 & 5.187 & 0.651 & 0.840 & 1.490 & 0.212 & 2.182 \\
\multirow{-3}{*}{SSL-Lanes~\cite{bhattacharyyassl}} & S/F Classification & 2.120 & 5.613 & 0.675 & 0.861 & 1.536 & 0.224 & 2.216 \\ \cmidrule(lr){2-2}
 & Scratch & 1.813 & 4.570 & 0.622 & 0.811 & 1.436 & 0.189 & 2.074 \\
\multirow{-2}{*}{Forecast-MAE} & Fine-tune w/o ensemble & \textbf{1.755} & \textbf{4.388} & \textbf{0.609} & \textbf{0.801} & \textbf{1.409} & \textbf{0.178} & \textbf{2.042} \\ \bottomrule
\end{tabular}
}
\vspace{1em}
\caption{Comparisons with previous results on the Argovesrse 2 test set (upper group) and validation set (lower group). For all the metrics, the lower is the better. We \textbf{bold} the best results and {\ul underline} the second best results.
}
\label{tab:leaderboard}
\vspace{-0.5em}
\end{table*}

\paragraph{Masking strategy.} 
One distinctive aspect of our approach, compared to existing SSL methods, is the introduction of agents' future trajectories as additional input during pre-training. The outcomes of various inputs and masking strategies are presented in Table \ref{fig:ab-mask_ratio}. When only lane masking is employed, the utilization of future trajectories as input makes a significant difference (minADE is 0.865 without using future, and 0.828 using future). One possible explanation is that the model can establish better connections between lanes and future trajectories through lane reconstruction, which is beneficial for the forecasting task. Interestingly, if we use the future as input and do not mask it, merely masking the history performs even worse than training from scratch (minADE 0.864/0.866 \vs 0.8314). A reasonable justification is that the dataset is intended to make the distribution of agents' future trajectories diverse and multi-modal (e.g., an agent is beginning to pass an intersection), while the historical trajectory is much simpler and more predictable. The model might take a shortcut to reconstruct the history by extrapolating the future trajectory, resulting in a failure to learn meaningful features from the agents' motion. As a result, the learned latent features are useless and even harmful for the later forecasting task. Adding future masking promptly addresses this problem, and minADE improves to 0.820 and 0.814. The proposed complementary masking strategy achieves the best performance in all metrics.

\paragraph{Encoder depth.} A relative deep encoder is necessary, as studied in Table \ref{tab:encoder_depth}. 
The performance improved 4.6\% in terms of minADE by increasing encoder depth from 2 to 4. 
Adding more encoder layers does not make a significant difference. 
We use an encoder depth of 4 as our default setting for its better efficiency-performance trade-off.

\begin{table}[t]
\centering
\tablestyle{8pt}{1.1}
\begin{tabular}{y{40}x{28}x{28}x{28}x{28}}
\toprule
\multicolumn{1}{l}{} & epochs & minADE$_6$ & minFDE$_6$ & MR$_6$ \\ \midrule
\multirow{3}{*}{Scratch} & 60 & 0.811 & 1.436 & 0.189 \\
 & 70 & 0.815 & 1.436 & 0.187 \\
 & 80 & 0.814 & 1.450 & 0.190 \\
 \hline
\baseline{Fine-tune} & \baseline{60} & \baseline{\textbf{0.801}} & \baseline{\textbf{1.409}} & \baseline{\textbf{0.178}} \\ \bottomrule
\end{tabular}
\vspace{1em}
\caption{\textbf{Comparison with training from scratch of different training epochs}. 
Continue increase training iterations does not further improves the performance of training from scratch. 
} 
\label{tab:more_epochs}
\vspace{-1em}
\end{table}

\subsection{Results}

For the final leaderboard submission, we use a depth of 4 for both the decoder and encoder. 
The history and lane masking ratios are 0.4 and 0.5, respectively. 
We set the pre-training and fine-tuning epochs both to 60. 
Our final motion forecasting model is simple and lightweight, with only 1.9M parameters in total.

\paragraph{Comparison with the other SSL method.}
We compare our method with SSL-Lanes, as it is the only published approach that employs vector representation and SSL. We make minimal modifications to its official code base\footnotemark~to adapt it to the AV2 dataset. Our experiments utilize three of its pretext tasks, specifically lane making, distance to the intersection (Dist. to Inter.), and success-failure classification (S/F classification). We do not implement the maneuver classification pretext task, as AV2 lacks lane-turning information. Table \ref{tab:leaderboard} (lower group) displays the comparison results on the AV2 validation set. Our Forecast-MAE outperforms all SSL-Lanes variants significantly across all metrics. Notably, SSL-Lanes suffers from performance degradation between the validation and test sets, whereas our approach achieves consistent performance on both sets and even performs slightly better on the test set. This suggests that our method learns superior and more generalized features through MAE-based self-supervised pre-training.

\footnotetext{\url{https://github.com/AutoVision-cloud/SSL-Lanes}}

\begin{figure*}[hbt!]
\begin{center}
\vspace{-0.8em}
\includegraphics[width=0.98\linewidth]{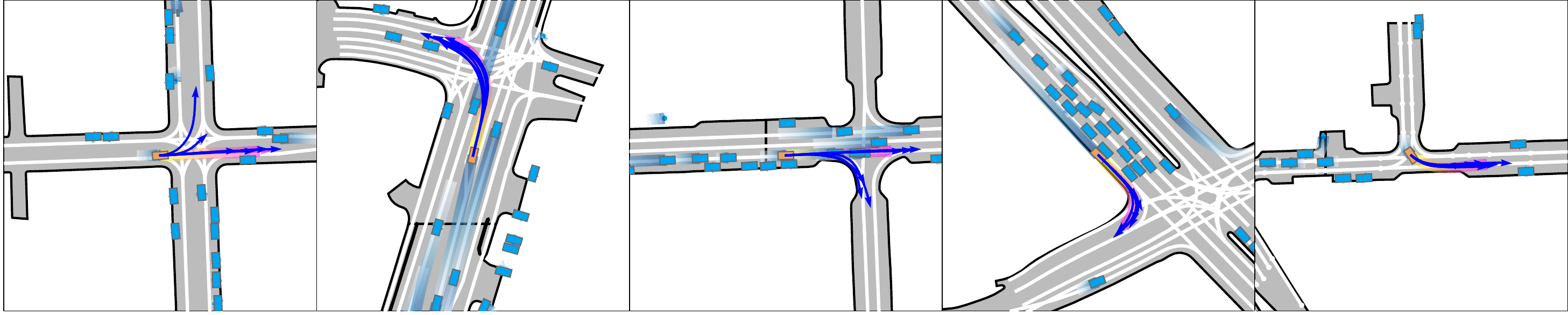}
\end{center}
\caption{\textbf{Qualitative results of Forecast-MAE} The predicted trajectories (K=6) are in blue and the ground truth is in gradient pink. The bounding box in orange indicates the focal agent.}
\vspace{-.8em}
\label{fig:visualization}
\end{figure*}

\paragraph{Comparison with State-of-the-art.} 
Our Forecast-MAE, developed using standard Transformer blocks and minimal prior knowledge, demonstrates impressive performance on the leaderboard, depicted in Table \ref{tab:leaderboard} upper group. Particularly noteworthy is that our approach (w/o ensemble) outperforms all other methods, including ensemble models, in terms of minADE$_1$ and minFDE$_1$, indicating its superior ability to predict the most likely future. We attribute this to our SSL pre-training scheme, which requires the model to reconstruct the most likely masked history and future trajectories. Additionally, Forecast-MAE (w/o ensemble) achieves the best minADE$_6$ among all non-ensemble methods and performs on par with QML (w/ ensemble). Through the adoption of an ensemble strategy involving 6 variants of our framework (\eg, different masking ratios, encoder depth), our ensemble model achieves the best performance among all methods across six metrics. In particular, our ensemble model outperforms the second-best (GANet) by 7.5\% in terms of minFDE$_1$.

\paragraph{Comparison with training from scratch.} 
The comparison results between the fine-tuned model and the model trained from scratch are presented in Table \ref{tab:leaderboard}. It is noteworthy that the vanilla model, despite its simplicity, serves as a strong baseline. However, our fine-tuned model outperforms the baseline in all metrics, exhibiting improvements of 5.1\% on minADE$_1$, 5.7\% on minFDE$_1$, 2.4\% on minADE$_6$, and minFDE$_6$, without the utilization of additional data or a more complex model.

As we incorporate agents' future trajectories as inputs during the pre-training, a plausible concern is that the fine-tuned model may benefit from additional training iterations. To address this, we conduct further training of the vanilla model with more epochs using cosine learning rate decay. The results presented in Table \ref{tab:more_epochs} indicate that continuing to increase the training iterations fails to enhance the performance of the model trained from scratch, underscoring the importance of pre-training.

\paragraph{Generalization ability.} 
Our method demonstrates strong generalization ability in the AV2 benchmark, as evidenced by the results in Table \ref{tab:leaderboard}. To further investigate this point, we design an experiment where training and testing employ different data distributions. Specifically, we partition all scenarios involving six cities in the AV2 dataset into two distinct and non-overlapping groups. We then train or pre-train the models solely on scenarios in \textit{Miami, Pittsburgh, and Austin}, and evaluate them on \textit{Dearborn, Palo-Alto, and Washington-DC}.  Results presented in Table \ref{tab:generalization} indicate that the fine-tuned model surpasses the baseline in all metrics, signifying that self-supervised pre-training enables learning of more generalizable features.

\paragraph{Qualitative Results.} We visualize the qualitative results of our fine-tuned model on the AV2 validation set, as shown in Figure \ref{fig:visualization}. 
We leave more results to the supplementary, due to the limited space.

\begin{table}[]
\tablestyle{12pt}{1.05}
\begin{tabular}{y{32}x{32}x{32}x{32}}
\toprule
 & minADE & minFDE & MR \\ \midrule
Scratch & 0.9098 & 1.645 & 0.2346 \\
Fine-tune & \textbf{0.8968} & \textbf{1.613} &\textbf{0.2164} \\ \bottomrule
\end{tabular}
\vspace{.8em}
\caption{\textbf{Evaluation results on different data distribution.}.
Models are trained or pre-trained on scenarios in \textit{Miami, Pittsburgh} and \textit{Austin} and validated in \textit{Dearborn, Palo-alto} and \textit{Washington-DC}. 
}
\label{tab:generalization}
\vspace{-1em}
\end{table}

\section{Conclusion}
We present Forecast-MAE, a simple and neat framework for self-supervised pre-training on the motion forecasting task. Based on the asymmetric architecture of MAE, we devise a scene reconstruction pretext task that utilizes a novel masking strategy. By leveraging the complementary masking of the agents' trajectories and the random masking of lane segments during the pre-training process, the model acquires the ability to capture the bidirectional agent motion features, road geometry features, and cross-modal interconnections jointly. Our experiments on the challenging Argoverse 2 benchmark demonstrate that our Forecast-MAE surpasses supervised learning methods and previous self-supervised learning works, especially in terms of
minADE$_1$ and minFDE$_1$, indicating its superior ability to predict the most likely future. 

\paragraph{Limitaion and Dicussion.}
One constraint of our work is the lack of exploration of transfer learning or few-shot learning for the proposed method (\eg, pre-training on WMOD and fine-tuning on AV2). 
Such exploration is hindered by the different problem settings, namely observation/prediction horizons, of different datasets. 
Besides, due to the relatively limited size of publicly available motion forecasting datasets compared to those in computer vision or natural language processing, we are unable to determine whether the performance of Forecast-MAE will scale up with increased training data and model capacity. 
However, we are positive about this point by drawing intuition from MAE and our minimal inductive bias design. 
Our approach could be advantageous for autonomous driving companies with large-scale internal datasets. 
Although Forecast-MAE already achieves strong performance while designed to be simple, we anticipate it can be further improved. 
Drawing inspiration from the development of techniques such as ViT~\cite{vit} to Swin-Trainsformer~\cite{liu2021swin}, properly incorporating inductive bias such as relative position design~\cite{zhou2022hivt, cui2022gorela, zhou2023query} or local attention~\cite{shimotion} may further boost Forecast-MAE in terms of performance and efficiency. 
Another possible direction is to generate realistic traffic scenarios building upon this work. 
These possibilities are left for future works. 

\paragraph{Acknowledgement} This work was supported by Guangdong Basic and Applied Basic Research Foundation, under project 2021B1515120032, Jiangsu Province Science and Technology Project BZ2021056, and Project of Hetao Shenzhen-Hong Kong Science and Technology Innovation Cooperation Zone (HZQB-KCZYB-2020083), awarded to Prof. Ming Liu.

{\small
\bibliographystyle{ieee_fullname}
\bibliography{egbib}
}

\clearpage
\newpage
\appendix

\section{Implementation Details}

\paragraph{Training.}
For all experiments, we train the model using an AdamW~\cite{loshchilov2017decoupled} Optimizer with a weight decay of 1e-4 and batch size of 128 on 4 GPUs. 
 We use cosine learning rate decay and the intial learning rate is 1e-3.
The dropout rate in all transformer blocks is set to 0.2. 
We use an agent-centric coordinates system and only consider agents and lane segments within 150 meters of the focal agent.  
The latent feature dimension is set to 128.
 
\begin{figure}
    \centering
    \includegraphics[width=0.95\linewidth]{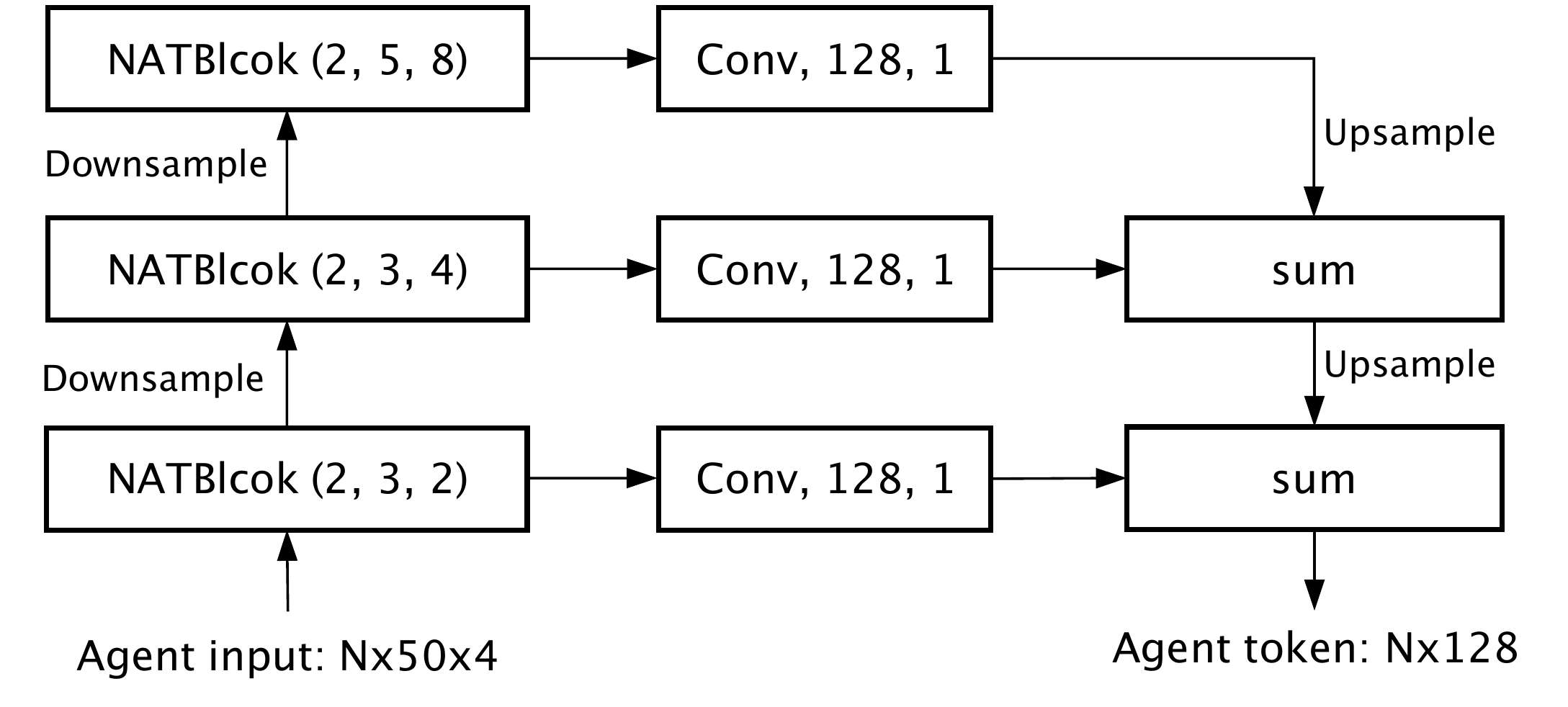}
    \caption{Detailed architecture of the agent history embedding layer. Numbers in the \textit{NATBlock} denote the number of stacked blocks, the kernel size of neighborhood attention, and the number of heads. Numbers in the \textit{Conv} indicate the hidden dimensions and stride.} 
    \label{fig:agent_embedding}
\end{figure}

\paragraph{Agent embedding.} 
The agent's embedding layer is a Feature Pyramid Network (FPN), primarily composed of neighborhood attention blocks (NATBlock) and 1D-convolution networks, depicted in Figure \ref{fig:agent_embedding}. The agent's input is of the shape $N\times50\times4$, which corresponds to a sequence of historical states spanning 5 seconds, sampled at a frequency of 10 Hz.
Each state includes the agents' displacement and velocity difference relative to the previous timestamp, along with a padding flag indicating the observation status.
The NATBlock exhibits an identical structure to the standard Transformer encoder block \cite{vaswani2017attention} (multi-head self-attention, add \& norm, and fully-connected layer), except for the replacement of self-attention with 1D neighborhood attention~\cite{hassani2022neighborhood}.
All downsample and upsample operators are implemented with 1D-convolution, having a down/up-sampling ratio of 2. We employ the same layer (but a seperate one) for the agents' future embedding during the pre-training phase.

\begin{figure}
    \centering
    \includegraphics[width=0.85\linewidth]{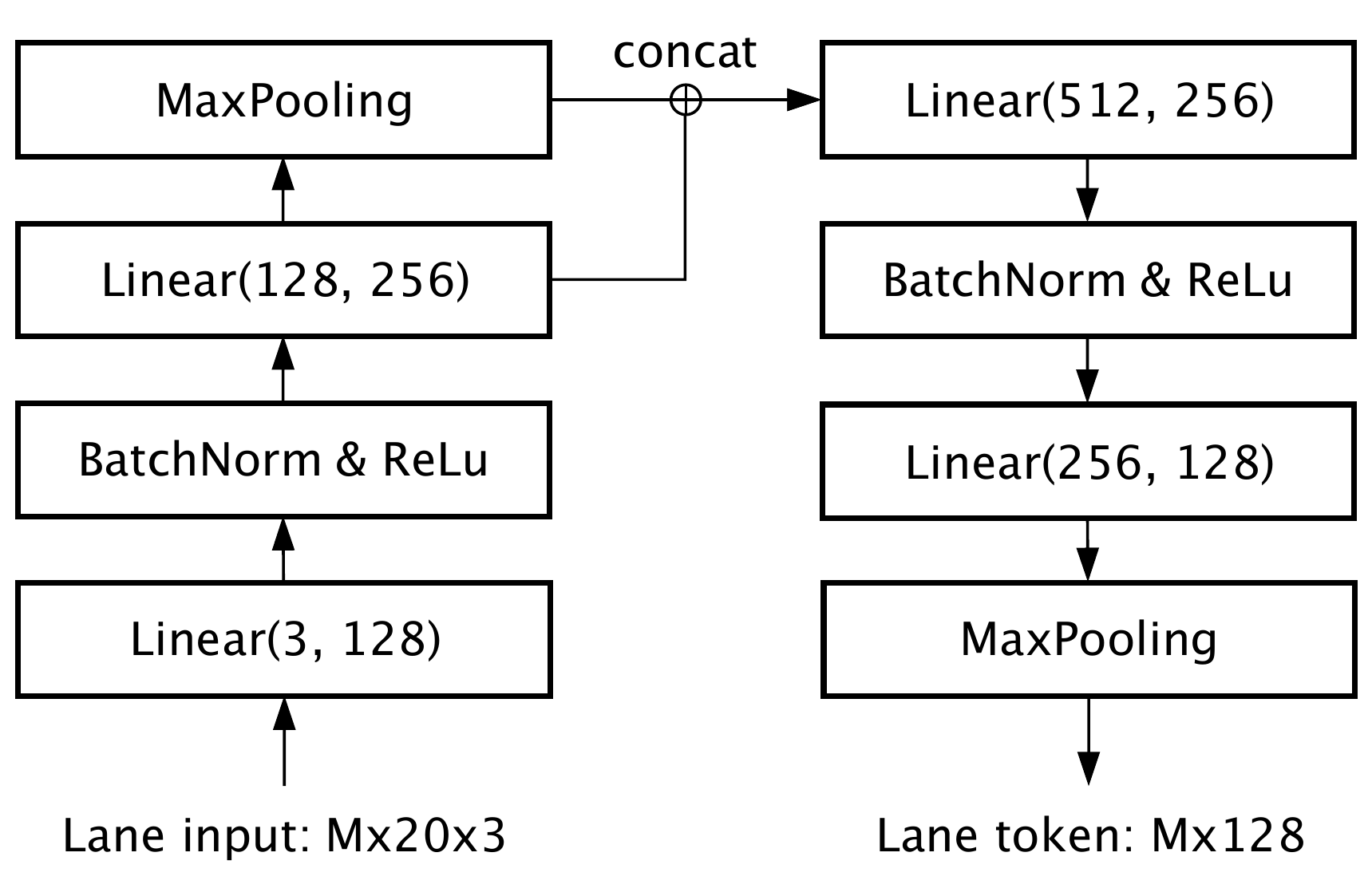}
    \caption{Detailed architecture of the lane embedding layer.}
    \label{fig:lane_embedding}
\end{figure}

\paragraph{Lane embedding.}  
The non-overlapping lane segments are acquired using the official Argoverse 2 API\footnotemark. Each individual lane segment is precisely interpolated to consist of 20 points. Each point encompasses its two-dimensional coordinates, normalized with respect to its geometrical center, as well as a padding flag denoting its presence within the region of interest to the focal agent. The architecture of the lane embedding layer adheres to the PointNet design~\cite{qi2017pointnet}, with a comprehensive depiction of its detailed structure provided in Figure \ref{fig:lane_embedding}.

\begin{figure}
    \centering
    \includegraphics[width=0.9\linewidth]{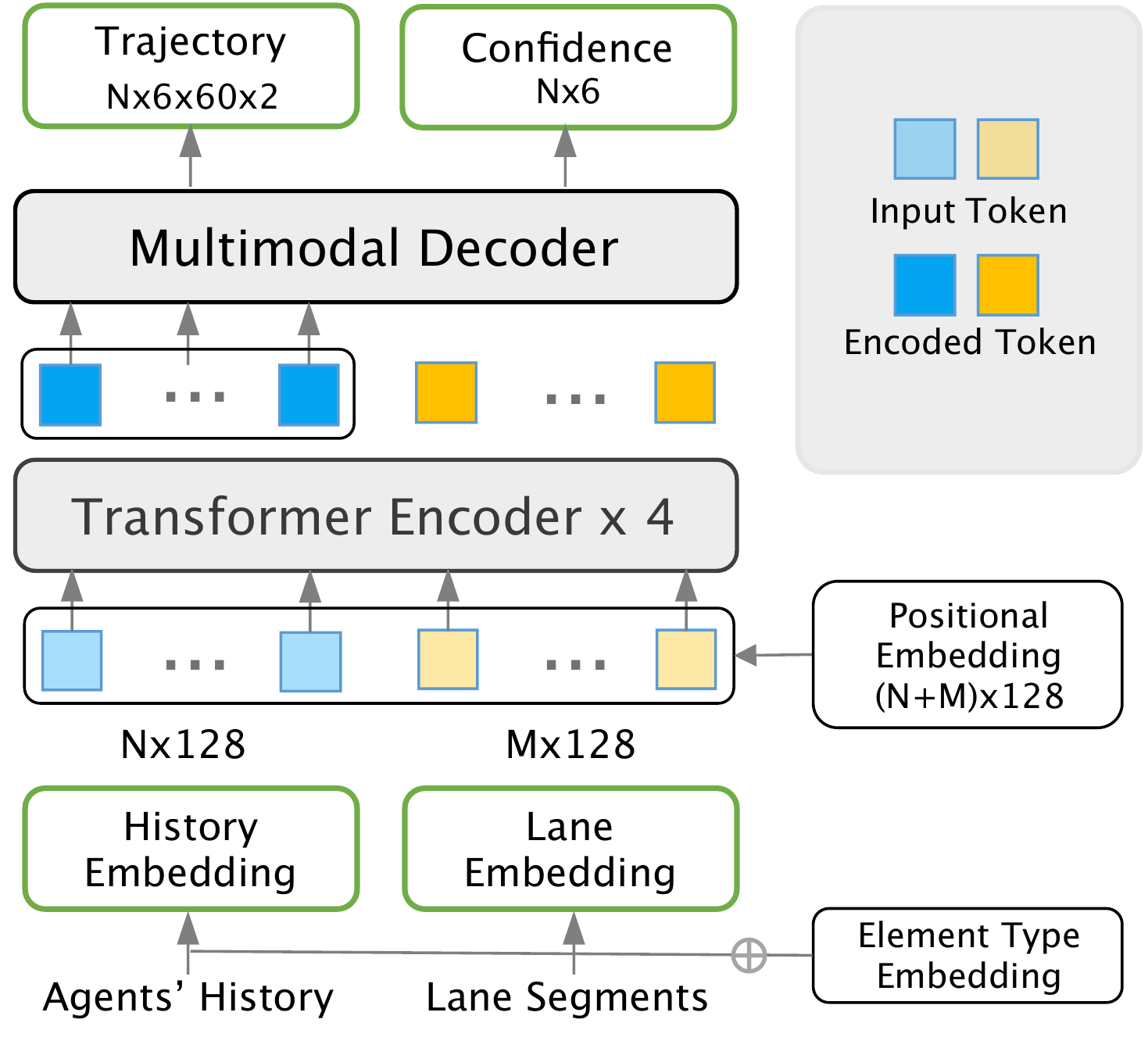}
    \caption{Overall architecture of fine-tune model}
    \label{fig:finetune}
\end{figure}

\paragraph{Fine-tune.}
We employ an end-to-end finetuning approach for the motion forecasting task, as illustrated by the overall architecture depicted in Figure \ref{fig:finetune}. 
Throughout the fine-tuning process,  only the history and lane features are embedded as inputs. 
Subsequently, the encoded history tokens of the agents are utilized for generating future predictions and associated confidences via the multi-modal decoder.

\paragraph{SSL-Lanes.}
To adapt SSL-Lanes~\cite{bhattacharyyassl} to the Argoverse 2 dataset, we change the history and future length to 50 and 60, respectively. 
The region of interest is increased from 100 to 150 meters accounting for the longer observation/prediction horizon. 
We follow the default experiment setting of SSL-Lanes, shown in Table \ref{tab:ssl_lanes}.

\footnotetext{\url{https://github.com/argoverse/av2-api}}

\paragraph{Experiment Setting.} We report the default setting for the pre-training and fine-tuning phase of Forecast-MAE in Table \ref{tab:pretraining} and Table \ref{tab:finetuning}.

\begin{table}[h]
\centering
\begin{tabular}{y{120}|x{60}}
config & value \\ 
\specialrule{.1em}{0.5ex}{0.5ex}
optimizer   &  AdamW     \\
learning rate   &  1e-3     \\
weight decay & 1e-4 \\
learning rate schedule   & cosine  \\
batch size & 128 \\
training epochs & 60 \\
warmup epochs & 10 \\
masking ratio & [0.4, 0.5] \\
loss weight & [1.0, 1.0, 0.35] \\
augmentation & none 
\end{tabular}
\vspace{1em}
\caption{Experiment setting for Forecast-MAE pre-training. Masking ratios refer to history trajectory and lane segments, respectively. }
\label{tab:pretraining}
\end{table}

\begin{table}[h]
\centering
\begin{tabular}{y{120}|x{60}}
config & value \\ 
\specialrule{.1em}{0.5ex}{0.5ex}
optimizer   &  AdamW     \\
learning rate   &  1e-3     \\
weight decay & 1e-4 \\
learning rate schedule   & cosine  \\
batch size & 128 \\
training epochs & 60 \\
warmup epochs & 10 \\
augmentation & none 
\end{tabular}
\vspace{1em}
\caption{Experiment setting for Forecast-MAE fine-tuning}
\label{tab:finetuning}
\end{table}

\begin{table}[h]
\centering
\begin{tabular}{y{120}|x{60}}
config & value \\ 
\specialrule{.1em}{0.5ex}{0.5ex}
optimizer   &  Adam     \\
learning rate   &  1e-3     \\
learning rate schedule   & 1e-4 at 62  \\
batch size & 128 \\
training epochs & 80 \\
augmentation & none 
\end{tabular}
\vspace{1em}
\caption{Experiment setting for SSL-Lanes.}
\label{tab:ssl_lanes}
\end{table}

\section{Additional Results}

\paragraph{Results on more datasets.}
We provide preliminary experimental results on Argoverse 1~\cite{chang2019argoverse} and WOMD~\cite{ettinger2021large}. The results are shown in Table \ref{tab:other_datasets}.

\begin{table}[t]
\tablestyle{8pt}{1.1}
\begin{tabular}{y{30}y{50}x{25}x{25}x{25}}
\toprule
\multicolumn{1}{l}{Dataset} & Method (year) & minADE & minFDE & MR \\ \midrule
\multirow{4}{*}{\begin{tabular}[c]{@{}c@{}}WOMD\\ (test set)\end{tabular}} & DenseTNT('21) & 1.039 & 1.551 & 0.157 \\
 & MTR('23) & 0.605 & 1.225 & 0.137 \\
 & Ours/scratch & 0.689 & 1.341 & 0.182 \\
 & \baseline{\textbf{Ours/fine-tune}} & \baseline{0.632} & \baseline{1.253} & \baseline{0.167} \\ \midrule
\multirow{4}{*}{\begin{tabular}[c]{@{}c@{}}AV1 \\ (val set)\end{tabular}} & TPCN('21) & 0.73 & 1.15 & 0.11  \\
 & Autobots('22) & 0.73 & 1.10 & 0.12 \\

 & Ours/scratch & 0.736 & 1.103 & 0.103 \\
 & \baseline{\textbf{Ours/fine-tune}} & \baseline{0.710} & \baseline{1.054} & \baseline{0.945} \\ 
 \bottomrule
\end{tabular}
\vspace{1em}
\caption{Preliminary results on WOMD test set and Argoverse 1 validation set, and comparison with representative methods. All results are from single model (w/o ensemble).}
\label{tab:other_datasets}
\end{table}

\begin{figure*}[hbt!]
\begin{center}
\includegraphics[width=0.88\linewidth]{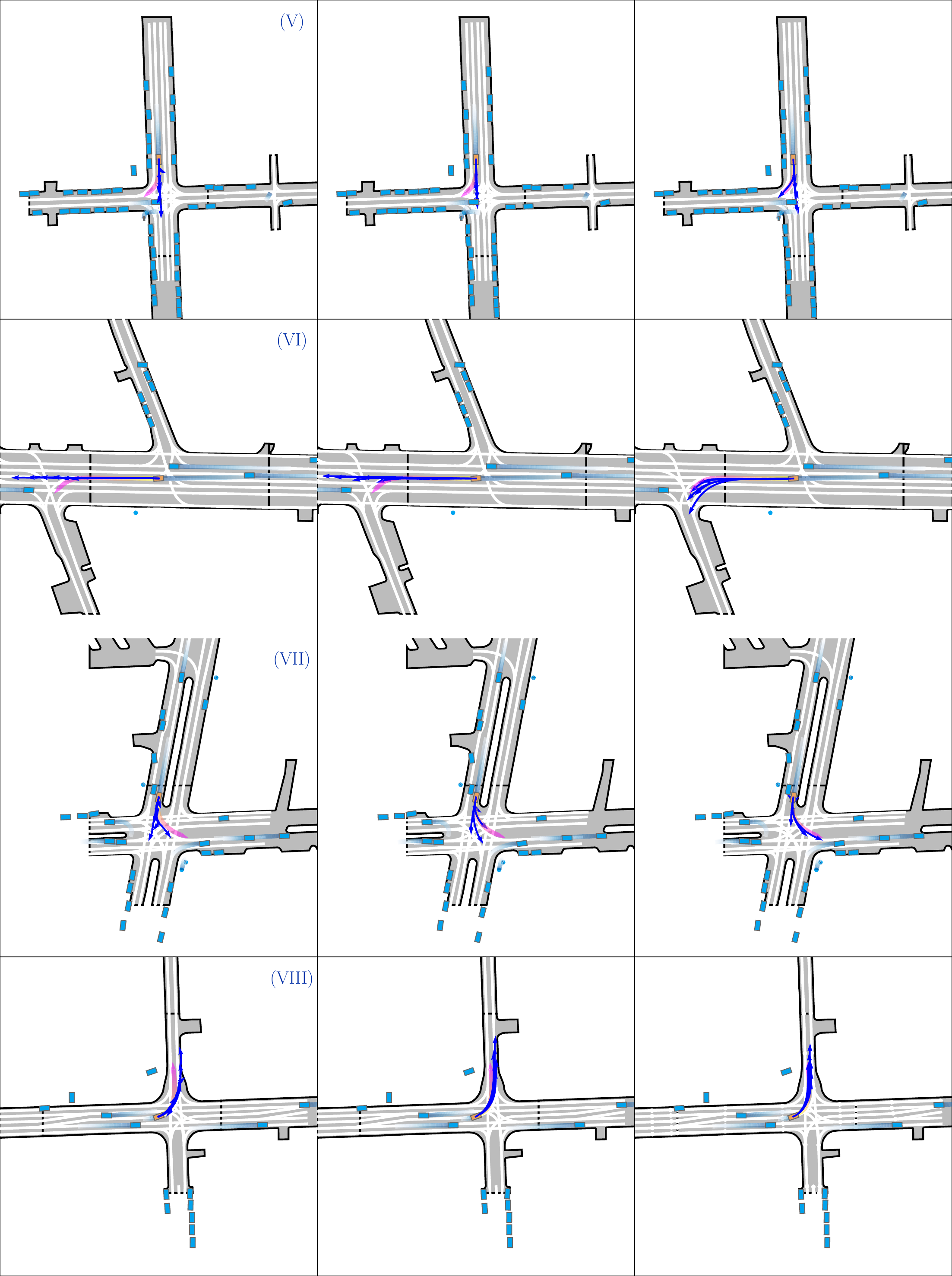}
\end{center}
   \caption{
   \textbf{More visualization results and comparisons on Argoverse 2 \textit{validation} set.} 
   The focal agent is denoted in orange, while the others are indicated in blue. The deep blue lines with arrows denote the predictions, and the gradual pink lines with arrows represent the ground truth. The arrows indicate the direction of motion. The gradual blue lines represent the historical trajectories, with the color transitioning from light to dark to indicate the direction of motion.
   }
\label{fig:visual}
\end{figure*}

\paragraph{More visual results and comparisons.}
We compare the performance of Forecast-MAE with two baselines, namely \textit{SSL-Lanes} and \textit{Scratch} (trained from scratch). The comparative visualization results are displayed in Figure \ref{fig:visual}. In comparison to the baselines, our Forecast-MAE model yields greater accuracy in direction and velocity prediction, even in high-speed and highly interactive scenarios. 
Notably, our fine-tuned model is the only one that captures lane-change behavior in scene (III). 
Furthermore, Forecast-MAE can generate a diverse range of multi-modal predictions while simultaneously ensuring precision, whereas other methods often predict infeasible trajectories. The visualization outcomes provide compelling evidence that our method is highly effective in encapsulating motion, road geometry, and cross-modal interaction features.

\begin{figure*}
    \centering
    \includegraphics[width=0.93\linewidth]{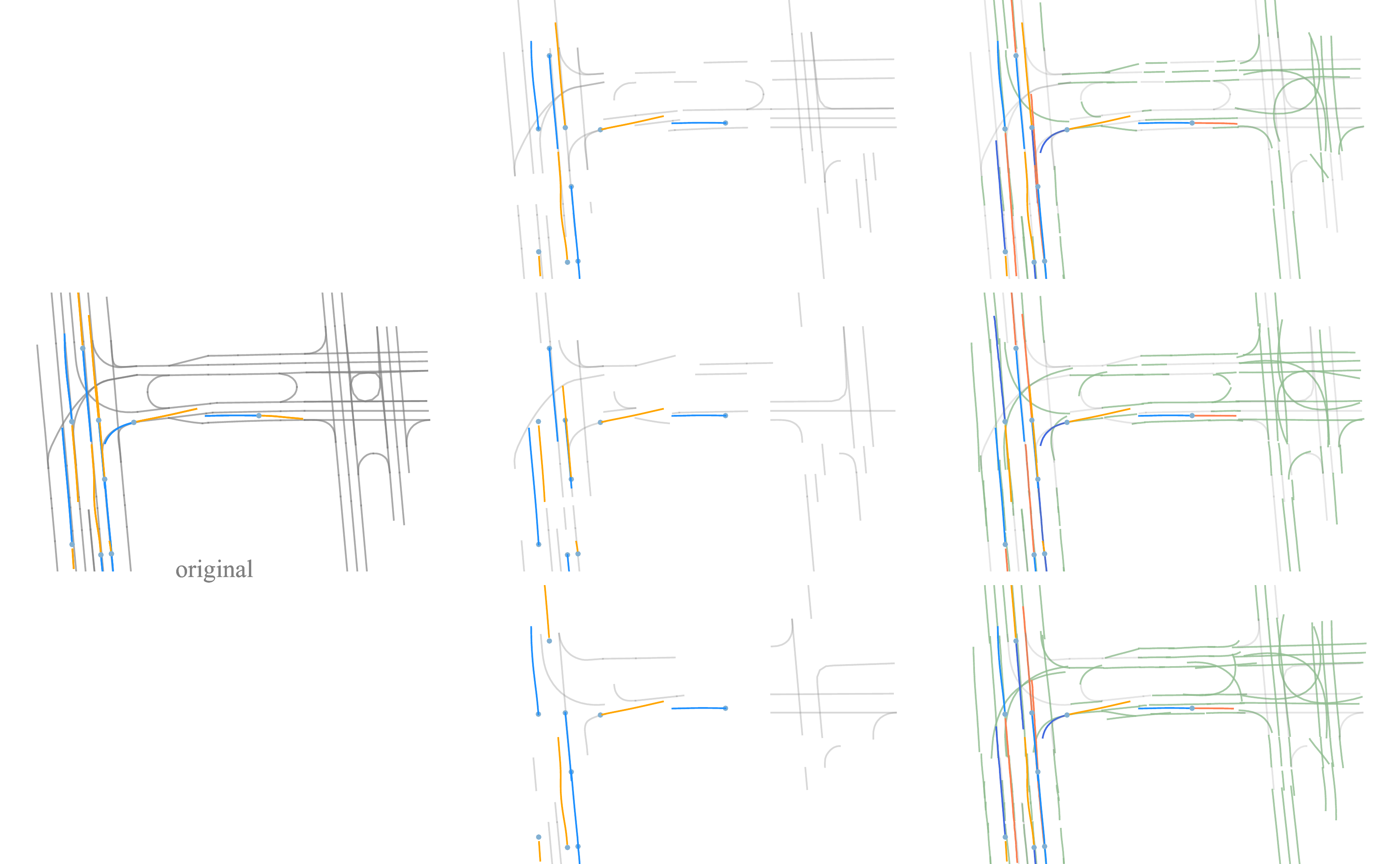}
    \caption{\textbf{Masked scene reconstructions results on Argoverse 2 \textit{validataion} set.} 
    The pre-trained model with a history/lane masking ratio of 0.5 is utilized to process input scenarios with higher lane masking ratios. The inputs and results are displayed in a top-down sequence, corresponding to the lane masking ratios of 0.5, 0.65, and 0.8.
    }
    \label{fig:recons}
\end{figure*}

\paragraph{Maksed scene reconstruction.}
We showcase the reconstruction results of two complex scenarios from Argoverse 2 validation set using our pre-trained model, which was trained with a history and lane masking ratio of 0.5. As depicted in Figure \ref{fig:recons} (first row), the pre-trained model exhibits a remarkable ability to recover the original scenario, including the history and future trajectories of agents and intricate lane geometries. 
Interestingly, our model performs well even with higher lane masking ratios (second and third rows in Figure \ref{fig:recons}). Despite a high lane masking ratio of 0.8, where most of the lane structures are lost, our model can still reconstruct most of the lane structures reasonably. These results suggest that our model has learned rich and profound scene representations through MAE-based self-supervised pre-training.

\begin{figure*}[hbt!]
\begin{center}
\includegraphics[width=0.88\linewidth]{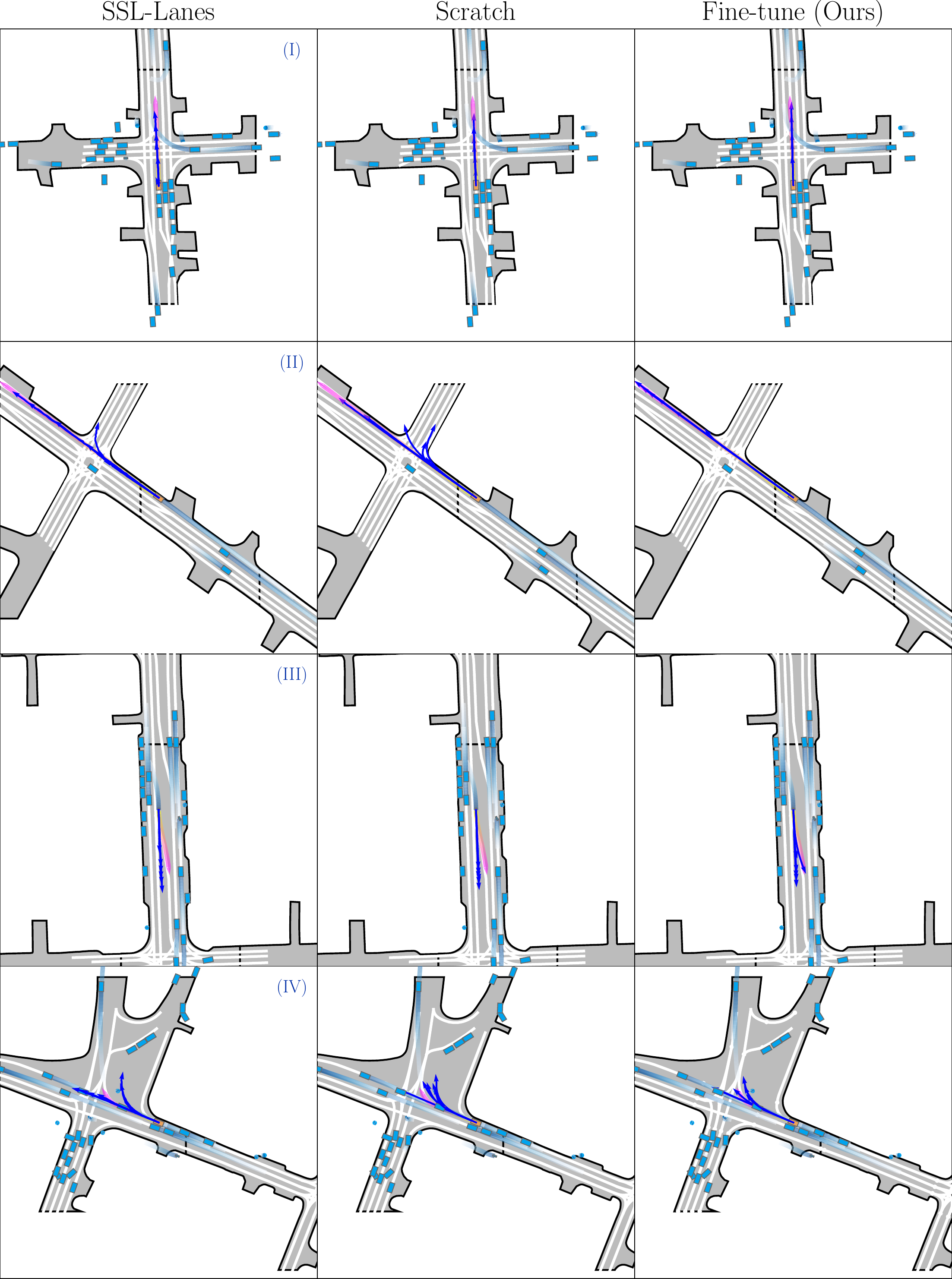}
\end{center}
\end{figure*}

\begin{figure*}[hbt!]
\begin{center}
    \includegraphics[width=0.93\linewidth]{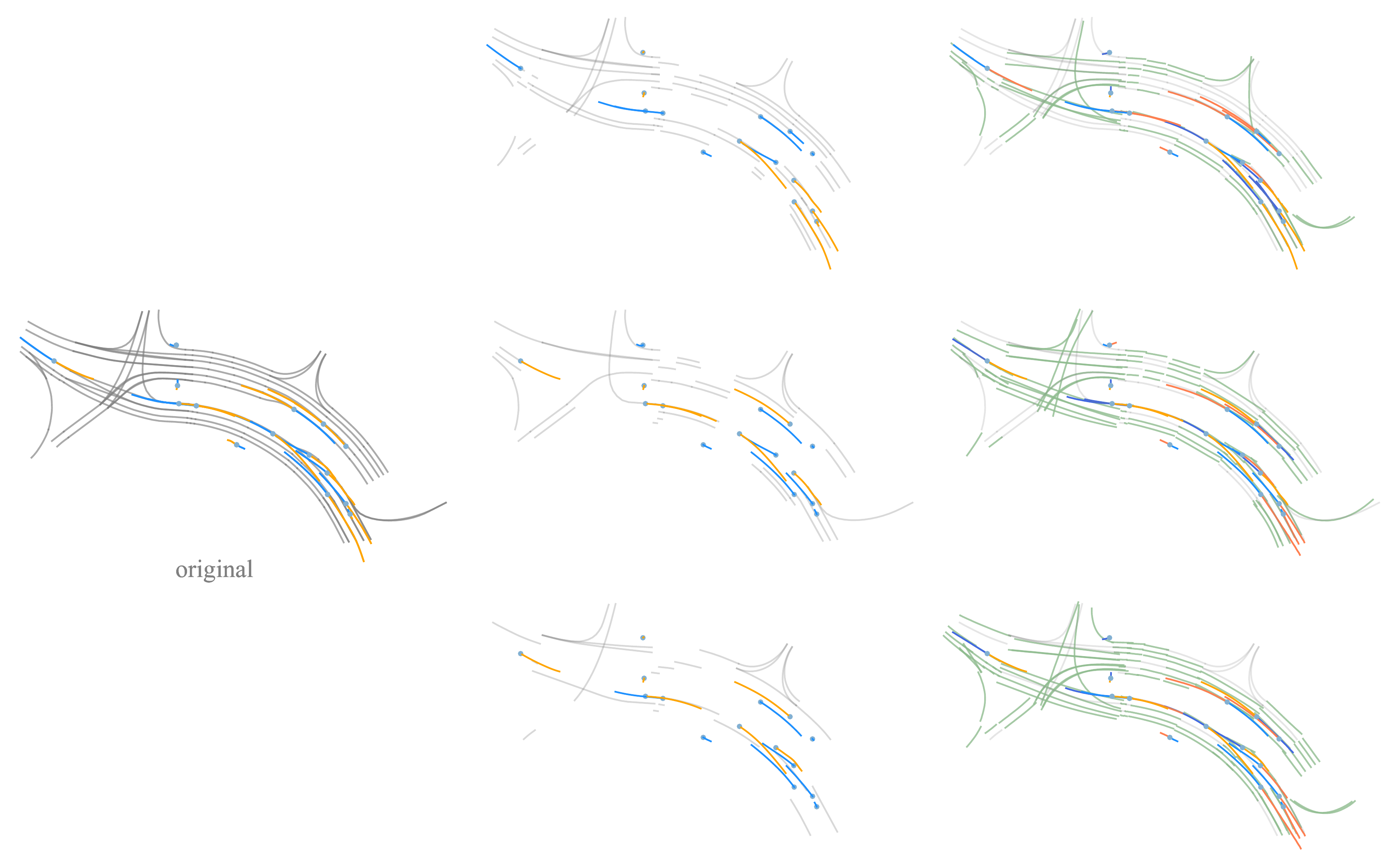}
\end{center}
\end{figure*}

\end{document}